\PassOptionsToPackage{table}{xcolor}
\documentclass[sigconf]{acmart}
\renewcommand\footnotetextcopyrightpermission[1]{} 
\usepackage{hyperref}
\usepackage{bm}
\usepackage{bbm}
\usepackage{tikz}
\usepackage{multirow}
\usepackage{arydshln}
\usepackage{subfiles}
\usepackage{subfig}
\usepackage{graphicx}
\usepackage{booktabs}
\usepackage{textcomp}
\usepackage{xspace}
\usepackage{stfloats}
\usepackage{float}
\usepackage{todonotes}
\usepackage[table]{xcolor}
\usepackage[linesnumbered,ruled,vlined]{algorithm2e}

\usepackage{footnote}
\makesavenoteenv{tabular}
\makesavenoteenv{table}

\acmConference[WWW]{WWW}{2021}{Slovenia}
\SetKwInput{KwInput}{Input}                
\SetKwInput{KwOutput}{Output}              
\SetKwFunction{FUpdateAndAggregate}{UpdateAndAggregate}

\definecolor{lgray}{gray}{0.95}
\newcommand{\commentout}[1]{}
\definecolor{cadmiumgreen}{rgb}{0.0, 0.42, 0.24}
\definecolor{darkgreen}{rgb}{0.3,0.5,0.3}
\definecolor{darkblue}{rgb}{0.3,0.3,0.5}
\definecolor{darkred}{rgb}{0.5,0.3,0.7}

\newif\ifshowcomments

\ifshowcomments
\newcommand{\mynote}[2]{\fbox{\bfseries\sffamily\scriptsize{#1}}
	{\small$\blacktriangleright$\textsf{\emph{#2}}$\blacktriangleleft$}}
\else
\newcommand{\mynote}[2]{}
\fi

\def\BibTeX{{\rm B\kern-.05em{\sc i\kern-.025em b}\kern-.08em
    T\kern-.1667em\lower.7ex\hbox{E}\kern-.125emX}}

\AtBeginDocument{%
  \providecommand\BibTeX{{%
    \normalfont B\kern-0.5em{\scshape i\kern-0.25em b}\kern-0.8em\TeX}}}

\newcommand{\alg}{\textsc{BiLA}\space}
\newcommand{\model}{\textsc{BiLA-CM}\space}

\newcommand{\modelapx}{\textsc{BiLA-WA}\space}

\author{
Chi Hong$^1$,
Amirmasoud Ghiassi$^1$,
Yichi Zhou$^2$,
Robert Birke$^3$,
Lydia Y. Chen$^1$,
\\ 
$^1$ Delft University of Technology,
$^2$ Tsinghua University,
$^3$ ABB Research, \\
\{C.Hong,S.Ghiassi\}@tudelft.nl,
zhouyc15@mails.tsinghua.edu.cn,
robert.birke@ch.abb.com,
lydiaychen@ieee.org
}

\begin{document}

\title{Online Label Aggregation: A Variational Bayesian Approach}

\begin{abstract}
Noisy labeled data is more a norm than a rarity for crowd sourced contents.
It is effective to distill noise and infer correct labels through aggregation results from crowd workers. To ensure the time relevance and overcome slow responses of workers, online label aggregation is increasingly requested, calling for solutions that can incrementally infer true label distribution via subsets of data items. 
In this paper, we propose a novel online label aggregation framework, \alg, which employs variational Bayesian inference method and designs a novel stochastic optimization scheme for incremental training.  \alg is flexible to accommodate any generating distribution of labels by the exact computation of its posterior distribution. We also derive the convergence bound of the proposed optimizer.
We compare \alg with the state of the art based on minimax entropy, neural networks and expectation maximization algorithms, on synthetic and real-world data sets. Our evaluation results on various online scenarios show that \alg can effectively infer the true labels, with an error rate reduction of at least 10 to 1.5 percent points for synthetic and real-world datasets, respectively. 
\end{abstract}


\maketitle

\section{Introduction}

\begin{figure*}[t]
\begin{center}
\includegraphics[width=1.7\columnwidth]{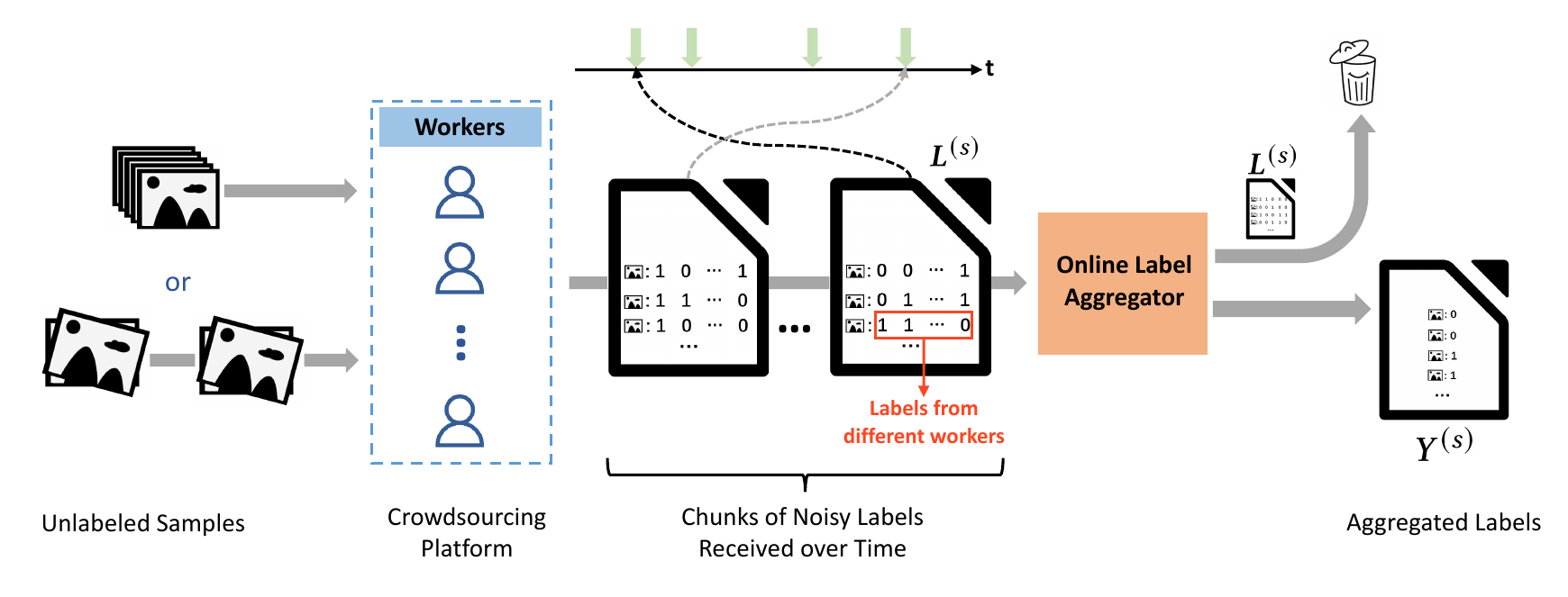}
\end{center}
\caption{The online label aggregation scenario.}
\label{fig:workflow}
\end{figure*}



Crowd sourcing platforms provide economic and efficient means to curate datasets which are deemed the new oil for today's artificial intelligence~\cite{imran2014aidr}. One of commonly seen crowd tasks is to classify contents, e.g., web pages~\cite{snow2008cheap}, and images~\cite{xiao2015learning}, and to provide labels of their respective classes. However, due to differences in the crowd workers' background and experience, the resulting labels of the same content often vary across workers, including missing labels -- so called noisy labels. The state of the practise~\cite{yang2018leveraging, yin2017aggregating} to distill the quality of crowd sourced labels is to aggregate them across all workers and reach consensus for every content. Such a curated dataset can then conveniently power up a wide range of supervised machine learning models for further analysis, e.g., object detection, search engine~\cite{snow2008cheap}, and disease diagnoses~\cite{heart1988}.

The velocity of knowledge discovery indeed hinges on the speed of data curation~\cite{Freitas2016}. Faster the data is aggregated via crowd sourcing, the more insights can be extracted through machine learning models. For example~\cite{lundgard2018bolt}, via instantaneous information from Amazon Mechanical Turk, i.e., in 2200 ms, the accuracy of predicting urban emergencies can be improved by 40\%. 
Moreover, labelling massive datasets come as a daunting tasks requiring months or years of effort. The estimated effort to label ImageNet for a single person working 24/7 is 19 years, but even with crowd sourcing involving 25K workers it still took 21 months~\cite{fei2010imagenet}.
It becomes increasingly imperative that label curation and aggregation can be conducted in online manner, i.e., labels can be continuously aggregated over a subset of content, instead of the entire content at once. More, recent privacy and governmental policies~\cite{EUdataregulations2018} regulate the data storage time, asking for prompt action of aggregation.

The key challenge behind online label aggregation is how to utilize a partial label set from workers that only includes a small chunk of content.
Existing aggregation methods~\cite{yin2017aggregating,dawid1979maximum,zhou2014aggregating} focus on the quality issues across workers but implicitly overlook the temporal aspect, i.e., timely and accurately label aggregation from online data. 
In other words, the prior art tailors for offline scenarios, which assumes the availability of all contents at once. As a result, in the online scenario, such approaches end up greedily optimizing for only the available subset, without the global optimization for the entire dataset. The need of online label aggregation thus calls for a novel stochastic optimization scheme which can handle batches of observable data.

Probabilistic graphic models~\cite{koller2009probabilistic} are commonly adopted to aggregate noisy labels from crowd workers without the label ground truths. Their objective is to maximize likelihood of the observed data by capturing the dependency on latent variables, e.g., the true labels and confusion matrix that specifies the generation process of label noise. Variational Bayesian inference methods~\cite{wainwright2008graphical,kurihara2007collapsed, teh2007collapsed} can effectively infer the latent features by maximizing the evidence lower bound~\cite{bishop2006pattern} of the log data likelihood of the observed data. 
The other popular approach to infer latent variables is Expectation-Maximization (EM) algorithm~\cite{dawid1979maximum} that has different objectives in expectation and maximization steps - an additional hurdle for stochastic optimization. 
While variational inference methods have an advantage of single objective for stochastic optimization, the challenges lies in deriving a tracktable posterior distribution of the generation process of label noise.

In this paper, we propose a novel online label aggregation framework, \alg, based on incremental variational Bayesian Inference method. \alg aggregates noisy labels from crowd workers incrementally upon receiving a subset of labeled items via a novel stochastic optimization scheme. To maximize the log likelihood of observed items, \alg minimizes the Kullback-Leibler (K-L) divergence between (i) the noisy label generative distribution $p$, and (ii) the approximate distribution $q$. The unique features of \alg are (i) flexibility and extendibility for generative distribution, (ii) exact computation of posterior distribution bypassing the need of the closed from expression, and (iii) the proposed objective function has the exact expression of the expectation term of K-L divergence, avoiding the approximation variance. 
Using the framework of \alg, we define a label aggregation model for multiple classes, abbreviated as \model, based on confusion matrix. We employ  multi-layer perceptron neural networks for approximate distribution $q$. 

As the data chunks are received in an online fashion, \model is incrementally trained by data chunks. To such an end, we propose a stochastic optimization scheme - a variant of RMSProp~\cite{Tieleman2012}. It enhances RMSProp with a dynamic clip operator, bias-corrected second raw moment estimate and decaying learning rate.

We evaluate \alg on both real-world and synthetic datasets. We compare its aggregation error rates with the state of the art label aggregation algorithms, i.e., Majority Voting, E-M based approaches, neural network based approaches and  Minimax Entropy based approaches. 
\alg is able to achieve significant error reduction in various online scenarios, i.e., different data chunk sizes. Our results also show that \alg is robust against different crowd sourcing scenarios, i.e., different number of workers, noise ratios, and label sparsity. In terms of effectiveness of proposed optimization scheme, we are able to achieve faster convergence than RMSProp, and in par with ADAM~\cite{kingma2014adam} but without risks of divergence. 

The contributions of this paper are summarized as follows.

\begin{itemize}
\item We design a flexible online label aggregation framework, \alg, based on variational Bayesian inference framework (\S~\ref{sec:method}). \alg uses neural networks for the approximate distribution guided by the generating distribution.
\item We provide a confusion matrix based aggregation model, \model, which outperforms existing algorithms based on EM algorithms, Minimax Entropy and neural networks (\S~\ref{sec:eval}).
\item We design a stochastic optimizer and derive its convergence bound (\S~\ref{sec:optimizer}). Last, we extensively compare \alg against representative label aggregation methods on different online crowd sourcing scenarios (\S~\ref{sec:eval})
\end{itemize}

\section{System Scenarios }
\begin{figure*}[t]
	\centering
	{
	\subfloat[Error rate after aggregating each chunk (chunk size:50 samples)]{
	    \label{subfig:offvson}
	    \includegraphics[width=0.49\textwidth]{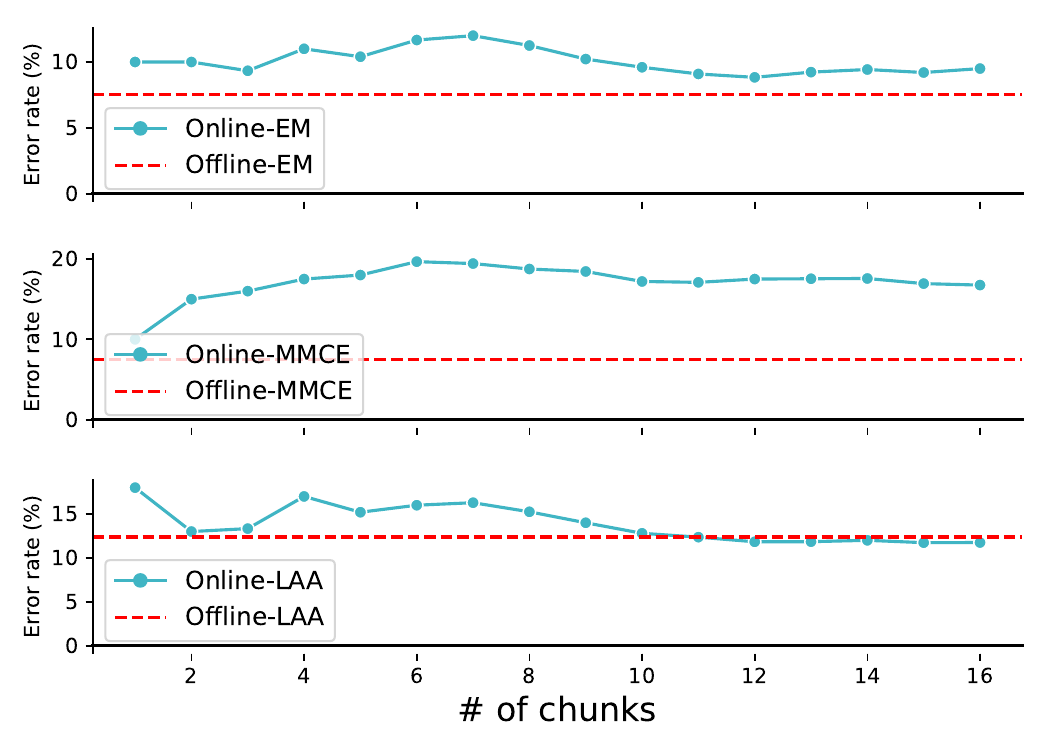}}
    \hfill
    \subfloat[The effect of the chunk size]{
	    \label{subfig:chunksize}
	    \includegraphics[width=0.49\textwidth]{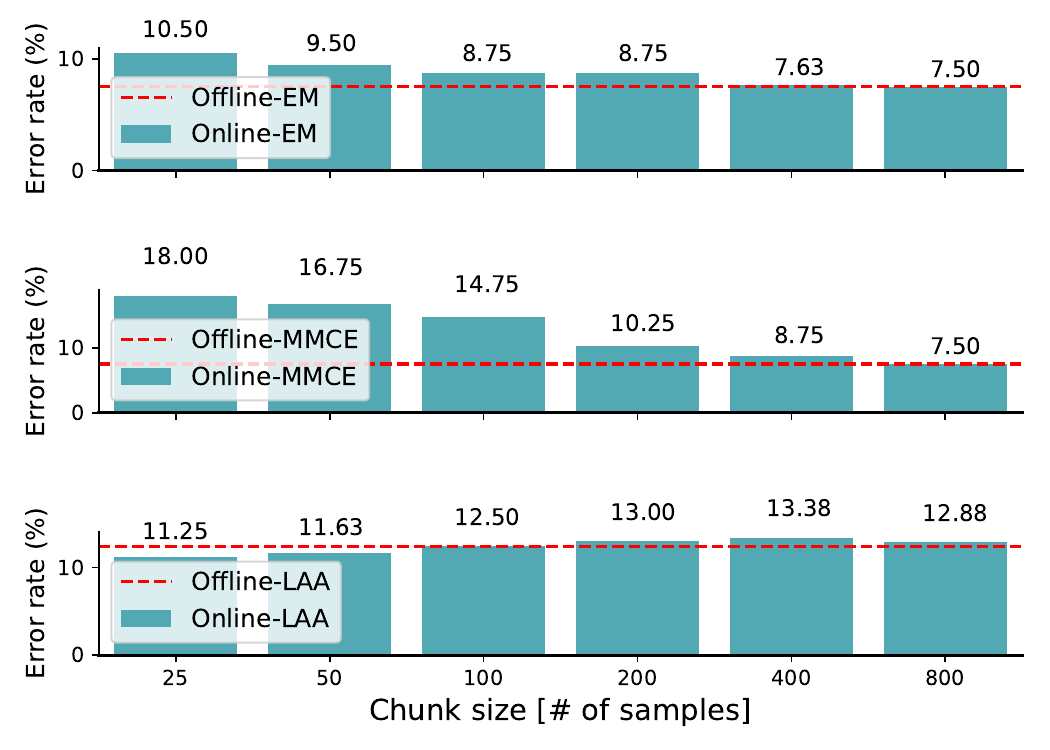}}
    }
	\caption{Motivation comparison on the RTE dataset.}
	\label{fig:motivation}
\end{figure*}

\label{system_scenarios}


To overcome the data labelling challenges it is common practise to label datasets via crowd sourcing by non-experts.
We can assign unlabeled instances to the workers in two ways: offline and online. In offline mode we publish all unlabeled instances once on a crowdsourcing platform and wait for all workers to complete their assignments before training the label aggregation algorithm. This works well if the data does not change over time and is all available at once.
In online mode, shown in Figure~\ref{fig:workflow}, we continuously publish single or multiple instances of unlabelled data on the crowdsourcing platform. Then we collect the labelling results and organize them in small chunks of redundant noisy labels. These chunks are fed one-by-one over time to the label aggregation algorithm to update the label aggregator.
The processed redundant noisy labels are discarded and only the aggregated, i.e. inferred true, labels are kept.
This enables continuous learning but requires the label aggregation algorithm to be be able to (incrementally) learn from small sets of data. This is challenging. Most state-of-the-art label aggregation techniques do not cope well with such a requirement.

We demonstrate this via a motivation example. We run three baseline aggregation algorithms in both online and offline mode and compare the achieved error rates. We consider Expectation-Maximization (EM)~\cite{dawid1979maximum}, Min-Max Conditional Entropy (MMCE)~\cite{zhou2014aggregating} and Label Aware Autoencoders (LAA)~\cite{yin2017aggregating} on the RTE dataset (details given in \S\ref{subsec:setup}). In online mode, 
we feed each aggregation algorithm with small chunks of 50 redundant noisy labels at a time. Each chunk is used to update the aggregator. We stop at 16 chunks (800 samples).
At each step we evaluate the achieved error rate. After each update we use the aggregator to infer the aggregated label for each sample and compute the percentage of samples for which the aggregated label differs from the ground truth label. Note that label aggregation is an unsupervised learning task. The ground truth labels are used only to compute the error rate, not to train the aggregator.
Figure~\ref{subfig:offvson} shows the step-wise error rate for the three methods.
Instead, Figure~\ref{subfig:chunksize} shows the sensibility of each method to the chunk size.
Each plot reports the achieved error rate when processing 800 samples in chunks of different size. For reference we report the offline performance, i.e. processing all 800 samples at once, as a horizontal line.

EM is commonly used to estimate a confusion matrix for each worker. MMCE is designed to discern the confusion matrices across the workers as well as the instances. Both can not be readily adapted to learn incrementally from small sets of data. EM uses majority voting results to determine a good starting point for the parameter search. Hence the best starting point is different for each chunk. MMCE assigns model parameters to each sample. Since each chunk has different samples, we can not keep the learned parameters.
We use this two methods in a sliding window style where each window is a new chunk of data.
As a result, EM and MMCE maximize the data likelihood of the current chunk not the full data.
Hence, the error rates of these two methods do not converge with time to the offline error rate, i.e. processing the whole data at once. More in detail, the performance of CE (see top plot Figure~\ref{subfig:offvson}) initially oscillates but then flattens out. After 16 chunks, i.e. all 800 samples, the error rate is still 2 percent points higher. MMCE is worse (see middel plot Figure~\ref{subfig:offvson}). The error rate first diverges before flattening out leaving a gap of 9.25 percent points after the last chunk. This is because MMCE is a generative model which needs to train a larger number of parameters compared to EM. This makes MMCE more sensible to the chunk size. This is clearly shown in Figure~\ref{subfig:chunksize}. The performance of both EM and MMCE (top and middle plot) benefit from processing larger chunks sizes. With chunks of 25 samples, EM is 3 percent point worse than offline, but starting at chunk size 400 EM is able to equal the offline performance. Instead, MMCE is more sensible. At chunk size 25 the gap is 11.5 percent points. The gap diminishes with increasing chunk sizes, but it never reaches the same performance as offline. Note that chunk size 800 is equivalent to offline.

LAA is a neural network based method inspired from autoencoders. This method can be used incrementally so that its optimization goal maximizes the data likelihood of the full dataset, not only the chunk. Consequently, online LAA nicely converges to the offline results over time (bottom plot Figure~\ref{subfig:offvson}). The small difference between the two is due to the stochasticity of the training. For the same reason the sensibility of this algorithm against different chunk sizes is low (bottom plot Figure~\ref{subfig:chunksize}). After processing 800 samples with different chunk sizes LAA achieves final error rates within $\pm 1$ percent points of the offline performance. However this method leads generally to worse results. The best result achieved by LAA is 11.75\% error rate compared to 7.5\% for EM and MMCE. LAA does not have a probabilistic model to describe the generative process of the observed noisy labels. This harms its performance. Besides LAA needs approximate approaches to calculate the expectation terms in the loss function. Our proposed label aggregation model \model directly addresses these issues achieving superior performance in both offline and online mode.

\section{Online Label Aggregation}
\label{sec:method}
We consider the online learning scenario shown in Figure~\ref{fig:workflow}. Each data instance $\bm{l}_i = \{l_{i0},...,l_{iK}\}$ contains redundant noisy labels of sample $i$. These noisy labels are provided by $K$ workers. $l_{ik} \in \bm{C}$ denotes the label of item $i$ given by worker $k \in \{1,...,K\}$, where $\bm{C} = \{1,...,C\}$ is the set of the possible classes. Not every worker might label all samples. If sample $i$ is not labeled by worker $k$, then the value of $l_{ik}$ is $-1$. We use $y_i \in \bm{C}$ to represent the unknown true label of a sample. Data instances stream into the label aggregation model at different times in small sets $L^{(s)}$. We call the small sets as chunks. In our online learning setting, our task is to continuously infer the values of $\{y_i| i \in L^{(s)} \}$ for the current chunk $L^{(s)}$ in real time before receiving the next chunk.

In order to define our optimization goal, we need some additional notations. We use $L$ to represent the collection of all observed noisy labels with $N$ instances, i.e. $\bm{L} = \{\bm{l}_{1}, ..., \bm{l}_{N}\}$, and 
$\bm{Y} = \{y_1,...,y_N\}$ for the collection of all the corresponding unknown true labels\footnote{To simplify notation we drop the subscript $i$ when referring to a generic sample.}.

\subsection{Variational Bayesian Inference Framework (\alg)}

\begin{figure}[t]
\begin{center}
\includegraphics[width=0.8\linewidth]{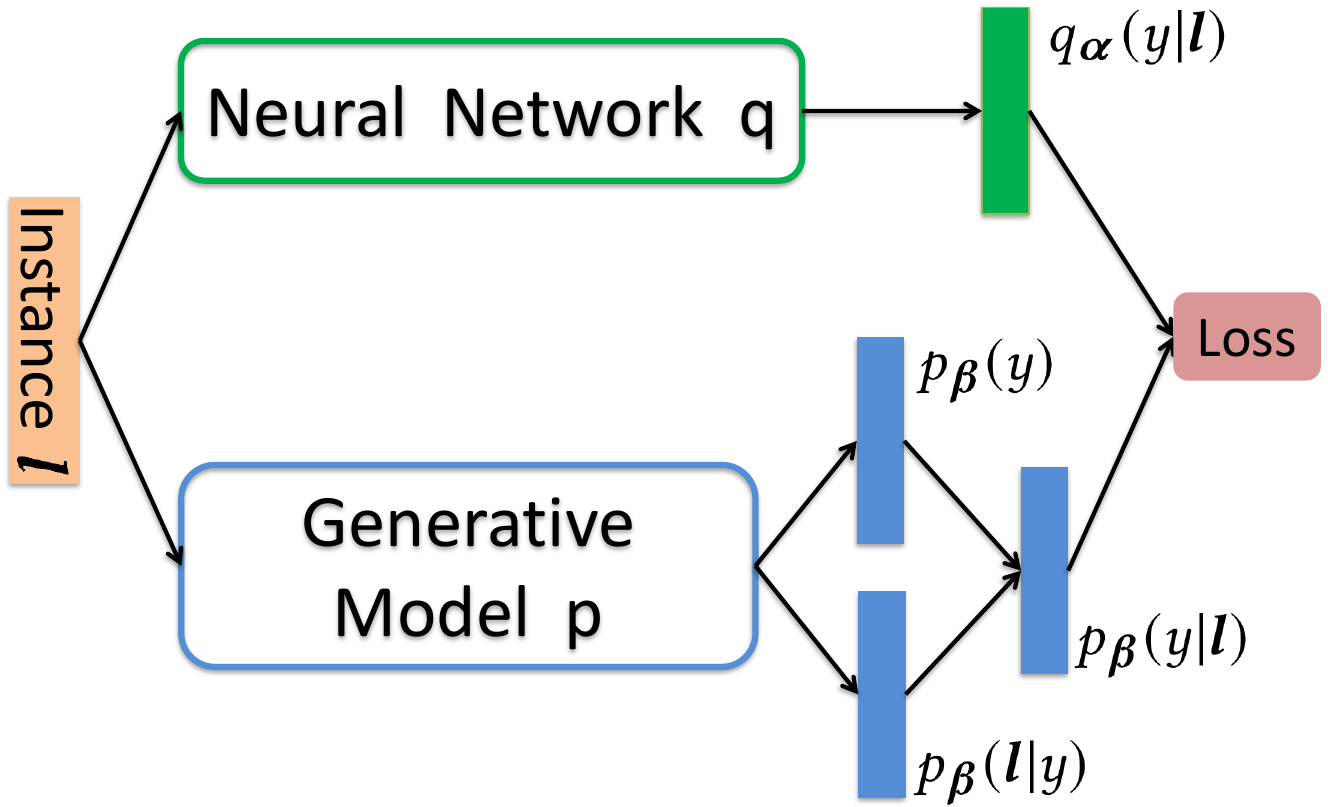}
\end{center}
\caption{The relationship between $q$, $p$ and the loss function.}
\label{alg_loss}
\end{figure}

In this section, we introduce our label aggregation framework (\alg) and define our optimization goal. The framework aims to predict the unknown true label $y_i$ of each instance $i$ with the sole knowledge of the instance's redundant noisy labels $\bm{l}_i$.
The framework includes two components: a neural network $q$ and a generative model $p$ trained using an optimization goal defined based on the principle of variational inference. From the perspective of variational Bayesian inference, $q$ is an approximate distribution. The choice of $q$ and $p$ is very flexible. $q$ can be a multilayer perceptron (MLP), a convolutional neural network (CNN), or any other neural network. $p$ is the model to define how to generate the observed noisy labels $\bm{L}$. Since we use a neural network as approximate distribution to learn the label aggregation model, we need stochastic optimization to train the parameters in $q$ and $p$. This requires that the loss function is differentiable respect to the model parameters in $p$. This is the only constraint of the definition of $p$. Besides, the close form of the posterior of model $p$ is needless because we rewrite the expression of the Kullback-Leibler divergence to avoid using the posterior directly. The relationship between $q$, $p$ and the loss function in \alg is shown in Figure~\ref{alg_loss}.

\subsubsection{Definition of $q$ and $p$}

The set of noisy labels $\bm{L} = \{\bm{l}_1,...,\bm{l}_N\}$ only contains the observed instances $\bm{l}_i$. The corresponding true labels $Y$ are unknown. The label aggregation task in this paper is to predict the unknown true labels given $\bm{L}$. So it is an unsupervised learning task.

Given an instance $\bm{l}_i$ of noisy redundant labels, we use a neural network $q$ with softmax activation on the last layer to predict the corresponding unknown true label $y_i$. $q$ can be represented as a probability distribution $q_{\bm{\alpha}}(y|\bm{l})$, where
$\bm{\alpha}$ denotes the neural network parameters. 
The output of the network is a C-dimensional vector $[q_{\bm{\alpha}}(y=c|\bm{l})]_{c=1}^{C}$, where the c-th element $q_{\bm{\alpha}}(y=c|\bm{l})$ is the probability that the true label of the input instance is class $c$. The predicted label is given by the element with the highest probability.

In order to train $q$ we need an optimization goal. Therefore, we define a generative model $p$ to describe the generative process behind the observed noisy labels $\bm{L}$. This way we can define a loss function to guide the training according to variational inference rules~\cite{bishop2006pattern,wainwright2008graphical}. $p$ assumes that an instance $\bm{l}$ is generated from some conditional distributions $p_{\bm{\beta}}(\bm{l}|y)$, where $y$ denotes the unknown true label and $\bm{\beta}$ the parameters of model $p$. It further assumes that $y$ is generated from a prior distribution $p_{\bm{\beta}}(y)$. The generative model $p$ potentially defines a posterior distribution:
\begin{equation}
p_{\bm{\beta}}(y|\bm{l}) = \frac{p_{\bm{\beta}}(\bm{l}|y)p_{\bm{\beta}}(y)}{p_{\bm{\beta}}(\bm{l})}.
\label{eq:gpos}
\end{equation}
We do not make many simplifying assumptions about $p$ except that the loss function is differentiable with respect to $\bm{\beta}$.

\subsubsection{Optimization Goal}

To solve the unsupervised learning task where we only have the observed noisy labels $\bm{L}$ a reasonable optimization goal is to maximize the data likelihood of $\bm{L}$. In particular, we maximize the data log likelihood $\log p_{\bm{\beta}}(\bm{L})$ according to the evidence lower bound $\log p_{\bm{\beta}}(\bm{L}) = KL(q_{\bm{\alpha}}(\bm{Y}|\bm{L})||p_{\bm{\beta}}(\bm{Y}|\bm{L})) + L(q) \ge L(q)$, where $KL(\cdot)$ denotes the Kullback-Leibler divergence and $L(q) = \mathbbm{E}_{q_{\bm{\alpha}}(y_i|\bm{l}_i)}\left[ \log \frac{p_{\bm{\beta}}(\bm{L}, \bm{Y})}{q_{\bm{\beta}}(\bm{Y})} \right]$. We can see that we can maximize the lower bound of $\log p_{\bm{\beta}}(\bm{L})$ as we minimize $KL(q_{\bm{\alpha}}(\bm{Y}|\bm{L})||p_{\bm{\beta}}(\bm{Y}|\bm{L}))$. So, we need to find the consensus between the predictions of $q$ and $p$. Consequently, we use $KL(q_{\bm{\alpha}}(\bm{Y}|\bm{L})||p_{\bm{\beta}}(\bm{Y}|\bm{L}))$ as our loss function and minimize it during the training process.

We assume that each collected label is generated independently, i.e. instances in $\bm{L}$ are independent from each other. Plugging $q_{\bm{\alpha}}(\bm{Y}|\bm{L}) = \prod_i q_{\bm{\alpha}}(y_i|\bm{l}_i)$ and $p_{\bm{\beta}}(\bm{Y}|\bm{L}) = \prod_i p_{\bm{\beta}}(y_i|\bm{l}_i)$ into the loss function, we have:
\begin{align}
KL(q_{\bm{\alpha}} & (\bm{Y}|\bm{L})||p_{\bm{\beta}}(\bm{Y}|\bm{L})) 
    = \sum_{i=1}^{N} -\mathbbm{E}_{q_{\bm{\alpha}}(y_i|\bm{l}_i)}\left[ \log \frac{p_{\bm{\beta}}(y_i|\bm{l}_i)}{q_{\bm{\alpha}}(y_i|\bm{l}_i)} \right] \notag \\
    &= \sum_{i=1}^{N} KL(q_{\bm{\alpha}}(y_i|\bm{l}_i)||p_{\bm{\beta}}(y_i|\bm{l}_i))
    \label{kl_sum}
\end{align}
Equation (\ref{kl_sum}) cannot be directly used to train $q$ and $p$, because the expression of $p_{\bm{\beta}}(y|\bm{l})$ is unknown. The exact expression of the posterior $p_{\bm{\beta}}(y|\bm{l})$ may be intractable. So we have to further rewrite the loss function. According to (\ref{eq:gpos}), we have
\begin{align}
&KL(q_{\bm{\alpha}} (y|\bm{l})||p_{\bm{\beta}}(y|\bm{l})) 
    = -\mathbbm{E}_{q_{\bm{\alpha}}(y|\bm{l})}\left[ \log \frac{p_{\bm{\beta}}(y|\bm{l})}{q_{\bm{\alpha}}(y|\bm{l})} \right] \notag \\
    &= -\mathbbm{E}_{q_{\bm{\alpha}}(y|\bm{l})}\left[ \log \frac{p_{\bm{\beta}}(y)}{q_{\bm{\alpha}}(y|\bm{l})} + \log p_{\bm{\beta}}(\bm{l}|y) \right] + const \notag \\
    &= KL(q_{\bm{\alpha}}(y|\bm{l})||p_{\bm{\beta}}(y)) - \mathbbm{E}_{q_{\bm{\alpha}}(y|\bm{l})}\left[ \log p_{\bm{\beta}}(\bm{l}|y) \right] + const
    \label{kl_calc}
\end{align}
To simplify the notations, we use $\theta = \{\bm{\alpha}, \bm{\beta}\}$ to represent the parameters of both models in our framework. According to (\ref{kl_sum}) and (\ref{kl_calc}) the loss function is rewritten as
\begin{align}
f(\theta ; \bm{L}) =& \frac{1}{N}\sum_{i=1}^{N} \{ KL(q_{\bm{\alpha}}(y_i|\bm{l}_i)||p_{\bm{\beta}}(y_i)) - \mathbbm{E}_{q_{\bm{\alpha}}(y_i|\bm{l}_i)}\left[ \log p_{\bm{\beta}}(\bm{l}_i|y_i) \right] \}
\label{eq:lossf}
\end{align}
where we ignore the constant term and the loss function is rescaled by $1/N$. This does not affect the optimization result.

\subsubsection{Training}

During training we solve the following optimization problem
\begin{equation*}
\hat{\theta} = \mathop{\arg\min}_{\theta} f(\theta; \bm{L}).
\end{equation*}
This optimization problem is solved by stochastic first-order optimization. The update rule of the model parameters is shown in Algorithm~\ref{alg:bilamc}. Following our discussions in  \S~\ref{system_scenarios}, in order to continuously aggregate small sets of noisy labels, we apply mini-batch training to update the parameters $\theta$. So the loss function for training is
\begin{align}
&f(\theta;\bm{L}^{(M)}) = \notag\\
&\frac{1}{M}\sum_{i=1}^{M} \left\{ KL(q_{\bm{\alpha}}(y_i|\bm{l}_i)||p_{\bm{\beta}}(y_i)) - \mathbbm{E}_{q_{\bm{\alpha}}(y_i|\bm{l}_i)}\left[ \log p_{\bm{\beta}}(\bm{l}_i|y_i) \right] \right\},
\label{eq:miniloss}
\end{align}
where $\bm{L}^{(M)}$ is a mini-batch sampled from current dataset $\bm{L}^{(s)}$, and $M$ denotes the minibatch size.
The gradient of $f(\theta; \bm{L}^{(M)})$ is required to update the model parameters. Before calculating the gradient, we need to define the expression of the KL divergence and the expectation term. The unobserved variables $y_i$ are discrete variables that take values from $1$ to $C$. Therefore, we have the following expressions
\begin{equation}
KL(q_{\bm{\alpha}}(y|\bm{l})||p_{\bm{\beta}}(y)) = -\sum_{c=1}^{C}q_{\bm{\alpha}}(c|\bm{l})\log \frac{p_{\bm{\beta}}(c)}{q_{\bm{\alpha}}(c|\bm{l})},
\label{eq:klcalc}
\end{equation}
\begin{equation}
\mathbbm{E}_{q_{\bm{\alpha}}(y|\bm{l})}\left[ \log p_{\bm{\beta}}(\bm{l}|y) \right] = \sum_{c=1}^{C}q_{\bm{\alpha}}(c|\bm{l})\log p_{\bm{\beta}}(\bm{l}|c),
\label{eq:ecalc}
\end{equation}
where $q_{\bm{\alpha}}(c|\bm{l})$ is the c-th element of the neural network output. Note that we do not require any approximation for calculating the expectation terms in our loss function. As such we avoid the problem of high variance of the loss function in stochastic Bayesian inference~\cite{paisley2012variational}.
According to (\ref{eq:klcalc}) and (\ref{eq:ecalc}), the values of $f(\theta;\bm{L}^{(M)})$ and corresponding stochastic gradient $\nabla_{\theta} f(\theta;\bm{L}^{(M)})$ can be easily computed. This completes all necessary blocks in the framework to construct online label aggregation models.

\subsection{Label Aggregation Model}

In this subsection, we introduce our online label aggregation model, \model,  based on the \alg framework. This model can be applied to aggregate discrete labels with noise.

\subsubsection{Model Definition}.
In order to define \model and exact loss function $f$, we need to decide the concrete forms of $q$ and $p$.
We set $q$ to be a fully connected neural network with a softmax activation function on the last layer. $q$ takes an instance $\bm{l}$ as input and outputs a distribution $q_{\bm{\alpha}}(y|\bm{l})$, where $\bm{\alpha}$ denotes the neural network parameters. 


$p$ is a generative model describing the observed noisy labels~$\bm{l}$. From (\ref{eq:klcalc}) and (\ref{eq:ecalc}), in order to compute the loss function and its gradient, we need to define the expressions of $p_{\bm{\beta}}(\bm{l}|y)$ and $p_{\bm{\beta}}(y)$. Since every element in an instance is collected independently from different workers, we assume that the $k$-th element in an instance is generated from an independent distribution $\bm{\psi}_{ck}$ when the true label of the instance is $c$. This distribution is defined as
\begin{equation}
\bm{\psi}_{ck} = softmax(\bm{\omega}_{ck}),
\label{eq:psi}
\end{equation}
where $\bm{\omega}_{ck}$ is a $C$-dimensional vector. Then $p_{\bm{\beta}}(\bm{l}|y = c)$ can be defined as
\begin{equation}
p_{\bm{\beta}}(\bm{l}_i|y_i = c) = \prod_{k \in \bm{S}_i}\psi_{ck,l_{ik}}, c \in [C],
\end{equation}
where $\psi_{ck,l_{ik}}$ is the $l_{ik}$-th element of $\bm{\psi}_{ck}$. Since the softmax function is derivable, $\bm{\omega}_{ck}$ can be updated by stochastic optimization. In this model, the prior distribution $p_{\bm{\beta}}(y)$ is a multinomial distribution estimated by
\begin{equation}
    \hat{p}_{\bm{\beta}}(y=c) = \frac{\sum_{i}\sum_{k}\mathbbm{I}(l_{ik}=c)}{\sum_{i}\sum_{k}\mathbbm{I}(l_{ik} \neq -1)}, c \in [C],
\label{eq:priorest}
\end{equation}
where the values of the estimators can be calculated by counting the observed labels. Since $p_{\bm{\beta}}(y)$ is fixed, we introduce a hyperparameter $\zeta$ to constrain the Kullback-Leibler divergence term in the loss function (\ref{eq:miniloss}). We regard this constrained term as a regularizer. Then, using (\ref{eq:klcalc}) and (\ref{eq:ecalc}) the mini-batch loss function used is
\begin{align}
f(\theta;\bm{L}^{(M)}) =& 
             -\frac{1}{M}\sum_{i=1}^{M} \Big\{ \zeta \sum_{c=1}^{C}q_{\bm{\alpha}}(c|\bm{l}_i)\log \frac{p_{\bm{\beta}}(c)}{q_{\bm{\alpha}}(c|\bm{l}_i)} \notag \\ &+ \sum_{c=1}^{C}q_{\bm{\alpha}}(c|\bm{l}_i)\log p_{\bm{\beta}}(\bm{l}_i|c) \Big\},
\label{eq:minilossprac}
\end{align}


\subsubsection{Online Model Training}.
The details of the complete online label aggregation model \model are illustrated in Algorithm~\ref{alg:bilamc}. \model continuously receives a new noisy labels chunk $\bm{L}^{(s)}$ containing multiple redundant noisy label instances $\bm{l}$. Note that we do not require the size of each set to be equal. This increases the practicality of our algorithm. At the beginning, we accumulate few noisy label sets to construct an initial set $\bm{L^{*}}$. This initial set is used to initialize the model parameters $\bm{\beta} = \{\bm{\omega}_{ck}\}$ and the prior estimator $\hat{p}_{\bm{\beta}}(y)$. $\bm{\beta}$ can be initialized by its definition and majority voting on the noisy redundant labels from the initial set to predict the true labels.
After initialization, we start the online aggregation. For each arriving chunk $\bm{L}^{(s)}$ at time step $t$, we update the model parameter $\theta$. Then we aggregate each $\bm{l} \in \bm{L}^{(s)}$ using the updated $\theta$. The \model update and aggregation process is illustrated in function \texttt{UpdateAndAggregate} (lines 10-30). First we retrain the model by computing the terms of the loss function $f$ from Equation (\ref{eq:minilossprac})  on each sampled mini batch (lines 14-21) before updating the model (lines 22-27).

\begin{algorithm}
\caption{Online Label Aggregation Model \model. The model parameters are $\theta = \{\bm{\alpha}, \bm{\beta}\}$, where $\bm{\alpha}=\{\bm{W}_1,\bm{W}_2,\bm{b}_1,\bm{b}_2\}$ and $\bm{\beta} = \{\bm{\omega}_{ck}\}$.}
\label{alg:bilamc}
\DontPrintSemicolon
  \textbf{Set:} learning rate $\mu > 0$, exponential decay rate $\gamma \in [0, 1)$, time step $t = 1$ \\
  \textbf{Input:} Continuously receive new noisy labels set $\bm{L}^{(s)} = \{\bm{l}\}$ \\
  Accumulate few sets to construct the initial set $\bm{L^{*}}$ \\
  Initialize $\theta$ using $\bm{L^{*}}$ \\
  $\bm{Y^{*}} = $ UpdateAndAggregate($\bm{L^{*}}$) \\
  \textbf{Output:} The aggregated labels $\bm{Y^{*}}$ \\
  \For{each arriving set $\bm{L}^{(s)}$}    
    { 
        $\bm{Y}^{(s)} = $ UpdateAndAggregate($\bm{L}^{(s)}$) \\
        \textbf{Output:} The aggregated labels $\bm{Y}^{(s)}$
    }
 
  \SetKwProg{Fn}{Function}{:}{}
  \Fn{\FUpdateAndAggregate{$\bm{L}^{(s)}$}}{
        \For{number of training epochs}
        {
            \For{number of minibatchs}
            {
                Sample a batch $\bm{L}^{(M)}=\{\bm{l}_1,...,\bm{l}_M \}$ from $\bm{L}^{(s)}$ \\
                \tcc{Calculate each term in $f$, Eq(\ref{eq:minilossprac})}
                \For{$c = 1,...,C$}
                {
                    \For{$k = 1,...,K$}
                    {
                        $\bm{\psi}_{ck} = softmax(\bm{\omega}_{ck})$
                    }
                }
                \For{instance $i = 1,...,M$}
                {
                    $\bm{h} = \bm{W}_2~tanh(\bm{W}_1\bm{l}_i + \bm{b}_1)+\bm{b}_2$ \\
                    $[q_{\bm{\alpha}}(y=c|\bm{l}_i)]_{c=1}^{C} = softmax(\bm{h})$ \\
                    \For{$c = 1,...,C$}
                    {
                        $\log g_{\bm{\beta}}(\bm{l}_i|y = c) = \sum_{k \in \bm{S}_i}\log \psi_{ck,l_{ik}}$
                    }
                }
                \tcc{update the model parameters $\theta$}
                $g_t \leftarrow \bigtriangledown_{\theta} f_t(\theta_t)$ \\
                $v_t \leftarrow \gamma \cdot v_{t-1} + (1-\gamma) \cdot (g_t \odot g_t)$ \\
                $\mu_t = \mu \cdot \sqrt{1-\gamma^{t}}$ \\
                $\eta_t = Clip(\mu_t/\sqrt{v_t}, \eta_{l}(t), \eta_{u}(t)) / \sqrt{t}$ \\
                $\theta_{t+1} \leftarrow \theta_t - \eta_t \odot g_t$ \\
                $t \leftarrow t+1$ \tcp*{count the time step}
            }
        }
        Get new confusion matrices $\bm{\pi}$ by the updated $\bm{\beta}$ \\
        Infer the aggregated labels $\bm{Y}^{(s)}$ by $\bm{\pi}$ \\
        \KwRet $\bm{Y}^{(s)}$
  }
\end{algorithm}

\subsubsection{Inferring the Aggregated Labels}.
Before introducing how to infer the aggregated label $y$, i.e. the predicted true label, for each sample $\bm{l}$, we discuss the connection between the generative model $p$ of \model and the confusion matrices of the workers.
The confusion matrix $\bm{\pi}_{c,z}^{(k)}$ of worker $k$ is a matrix for describing the worker's labeling behavior~\cite{dawid1979maximum}. The matrix element $\pi_{c,z}^{(k)} = p(l_{ik} = z | y_i =c)$ is the probability that worker $k$ assigns the label $z$ to the instance $i$ when the true label $y_i$ is $c$. According to the definition of $p$, we have that $\psi_{ck,z} = p_{\bm{\beta}}(l_{ik} = z| y_i=c)$ which corresponds to $\pi_{c,z}^{(k)}$. Therefore, we can easily construct the confusion matrices of the workers after learning the parameters $\bm{\beta}=\{\bm{\omega_{ck}}\}$. Note that this provides insight on the noise process which other methods lack, e.g. LAA. With the confusion matrices the inference problem becomes trivial. After obtaining the values of the confusion matrices, we can infer the aggregated label of an instance by maximizing the data likelihood of the corresponding observed noisy labels, where $p(\bm{l}_i|y_i=c,\bm{\pi}) = \prod_{k=1}^{K}\prod_{z=1}^{C}(\pi_{c,z}^{(k)})^{\mathbbm{I}(L_{i,k}=z)}$. $\mathbbm{I}(\cdot)$ is an indicator function taking the value $1$ when the predicate is true, and $0$ otherwise,

\section{Optimizer and Convergence Analysis}
\label{sec:optimizer}
We propose a stochastic optimizer to train   \model and summarize key steps in line 22-27 of Algorithm~\ref{alg:bilamc}. It's a variant of RMSProp~\cite{Tieleman2012}. The update of the model parameter $\theta$ (line 26) is based on gradient rather than the momentum. Similar to  RMSProp, we utilize a second raw moment estimate of the gradient (line 23) to obtain the element-wise adaptive learning rates for every element of $\theta$ (line 24-25). The element-wise adaptive learning rates are important because of the observation that in a multilayer neural network, the appropriate learning rates can vary widely between weights~\cite{Tieleman2012}. Furthermore, we apply a clip operator applied to avoid gradient explosion (line 25). In order to avoid an abrupt stop in training, we employ decayed learning rate approach. The upper and lower bound of the clip operator is then divided by $\sqrt{t}$ to obtain decayed element-wise learning rates. 

Furthermore, we also analyze the convergence property of our stochastic optimization approach in the online convex framework~\cite{zinkevich2003online}. According to the framework setting, we use an unknown sequence of convex loss functions $f_{1}(\theta), f_{2}(\theta), ..., f_{T}(\theta)$ to represent the loss functions at each iteration time step $t$. In each time step the training data (mini-batches) are different, so we need different notations to represent the stochasticity of the loss functions. The regret is applied to evaluate the convergence of our label aggregation algorithm. The regret is defined as $R(T) = \sum_{t=1}^{T}[f_t(\theta_t)-f_t(\theta^{*})]$ where $\theta^{*} = \mathop{\arg\min}_{\theta \in \chi} \sum_{t=1}^{T} f_t(\theta)$. Actually $R(T)$ represents the sum of the difference between the online prediction $\theta_t$ and the best fixed parameter $\theta^{*}$. We will show that our algorithm has a $O(\sqrt{T})$ regret bound. The details of the derivation process is shown in the Appendix~\ref{Appendix_Bound}. To show the bound, we define the notation $\eta_{t,i}$ to be the $i^{th}$ element of $\eta_{t}$.

~\\
\noindent \textbf{Theorem 1.} \emph{Let $\{\theta_{t}\}$ be the parameter sequence obtained from our optimizer where $\theta \in R^{d}$. Suppose $\eta_{u}(t) \le R_{\infty}$ and $\frac{t}{\eta_{l}(t)} - \frac{t-1}{\eta_{u}(t-1)} \le B$ for all $t \in [T]$. Assume that $\left\| \theta_n - \theta_m \right\|_{\infty} \le D_{\infty}$ for all $\theta_n, \theta_m \in \chi$ and $\left\| \bigtriangledown f_t(\theta) \right\|_{2} \le G$ for all $t \in [T]$ and $\theta \in \chi$. Our optimizer have the following guarantee of the regret}
\begin{equation*}
    R(T) \le \frac{1}{2}D_{\infty}^{2}\left[ 2dB(\sqrt{T}-1) + \sum_{i=1}^{d} \eta_{1,i}^{-1} \right] + (\sqrt{T}-\frac{1}{2})R_{\infty}G^{2}.
\end{equation*}
~\\

According to Theorem~1, if we choose the lower and the upper bounds of the clip operator which satisfy $\eta_{u}(t) \le R_{\infty}$ and $\frac{t}{\eta_{l}(t)} - \frac{t-1}{\eta_{u}(t-1)} \le B$ for all $t \in [T]$, the regret bound will be $O(\sqrt{T})$. The lower bound and upper bound with constant values definitely satisfy these conditions. Dynamic clip bounds as shown in the experimental part of the paper~\cite{savarese2019convergence} also meet the requirement and they have good performance in practise. We choose these bounds to conduct our experiments. Note that because of the clip operator, $\sum_{i=1}^{d} \eta_{1,i}^{-1}$ takes a limited value. Then we have the corresponding convergence rate $\frac{R(T)}{T} = O(\frac{1}{\sqrt{T}})$ where $\lim_{T \to \infty} \frac{R(T)}{T} = 0$. That shows the average of the difference between the online prediction and the best fixed parameter tend to $0$ during the iterations. Thus, the regret bound guarantees the convergence of our algorithm.

\section{Evaluation}
\label{sec:eval}
\subsection{Experimental setup}
\label{subsec:setup}
\subsubsection{Datasets}

\begin{table}[t]
\centering
\caption{Datasets overview.}
\label{tab:datasets}
\begin{tabular}{|c|c|c|c|c|} 
\hline
\rowcolor{gray!25}
\textbf{Dataset} & \textbf{Workers} & \textbf{Items} & \textbf{Labels} & \textbf{Classes} \\ 
\hline
\hline
\textit{Adult}   & 17 & 263 & 1370 & 4 \\ 
\hline
\textit{RTE}     & 164 & 800 & 8000 & 4 \\ 
\hline
\textit{Heart}   &  12 & 237 & 952 & 2 \\ 
\hline
\textit{Age}     &  165 & 1002 & 10020 & 7 \\ 
\hline
\textit{CIFAR10$^S$}   &  10 & 50K & 45K & 10 \\
\hline
\textit{Pendigits$^S$}   &  10 & 11K & 9.9K & 10 \\
\hline
\end{tabular} \\
\footnotesize{
$^S$ uses synthetic redundant noisy labels.}
\end{table}

We consider six different datasets in our experiments to evaluate the performance of \model comparing to competitors. In our experiments, CIFAR-10 and Pendigits are the only synthetic dataset, while the rest of them are real-world datasets.

\begin{itemize}
    \item \textbf{Adult}~\cite{ipeirotis2010quality}: It contains data labeled by Amazon Mechanical Turk workers. The labels are categorized into four classes based on the amount of adult content on each web page. 
    \item \textbf{RTE}~\cite{snow2008cheap}: It includes 164 workers for assigning labels of 800 items into 2 classes of textual entailment. 
    \item \textbf{Heart}~\cite{heart1988}: It is a dataset provided by 12 medical students categorizing the patients into 2 groups of heart and non-heart diseases based on physical examination. It has 12 workers for 237 samples. 
    \item \textbf{Age}~\cite{han2015demographic}: This dataset is the 1001 faces of different people who have been labeled with their age. In our experiments, the labels are discretized into 7 age groups: [0,9], [10,19], [20,29], [30,39], [40,49], [50,59], [60,100].
    \item \textbf{CIFAR-10}~\cite{kriz-cifar10}: It is a vision dataset including 50K $32 \times 32$-pixels training images classified into 10 classes. Here we create synthetic noisy labels. For each image we generate $[6, 8, 10]$ redundant labels, i.e. workers, drawn from a bimodal noise distribution with $[0.4, 0.6, 0.8]$ mislabelling probability and $[0.1, 0.2, 0.3]$ missing label probability. We center the bimodal distribution around classes $\mu_1=3.0$, $\mu_2=7.0$ with variance $\sigma_1=1.0$, $\sigma_2=0.5$.
    \item \textbf{Pendigits}~\cite{Dua:2019}: This dataset targets the recognition of 10992 handwritten digits from 44 writers. We create synthetic redundant noisy labels using the same noise model as for CIFAR10.
\end{itemize}
Table~\ref{tab:datasets} summarizes the characteristics of all datasets.

\begin{table*}
\centering
\caption{Online  label aggregation: error-rates (\%). The online version of baseline algorithms are appended with prefix of "o". }
\label{tab:online_error-rates}
\resizebox{2\columnwidth}{!}{%
\begin{tabular}{|c|c;{1pt/1pt}c;{1pt/1pt}c;{1pt/1pt}c;{1pt/1pt}c|c;{1pt/1pt}c;{1pt/1pt}c;{1pt/1pt}c;{1pt/1pt}c|} 
\hline
\rowcolor{gray!25}      & \multicolumn{5}{c|}{Small chunk size}                                                                  & \multicolumn{5}{c|}{Big chunk size}                                                                   \\ 
\cline{2-11}
\rowcolor{gray!25} \multirow{-2}{*}{\textbf{Dataset} } & \textbf{MV} & \textbf{oEM} & \textbf{oMMCE} & \textbf{oLAA} & \textbf{\model} & \textbf{MV} & \textbf{oEM} & \textbf{oMMCE} & \textbf{oLAA} & \textbf{\model}  \\ 
\hline
\hline
\textit{CIFAR10 (200, 500)}                                                                                   &            24.74              &          22.58           &           43.00             &      29.20        &       \textbf{13.29}         &       24.74                   &         17.68            &          14.25             &     30.29         &          \textbf{13.69}       \\ 
\hline
\textit{Pendigits (200, 500)}                                                                                        &             25.27             &           23.11          &           44.77             &       28.84       &        \textbf{13.34}        &             25.27             &          18.15           &           14.77            &       30.54       &       \textbf{13.06}          \\ 
\hline
\textit{Age (25, 50)}                                                                                        &           34.73               &          35.03           &             41.52          &      35.53        &       \textbf{33.73}         &                34.73          &         34.43            &          41.92             &       36.33       &         \textbf{33.63}        \\ 
\hline
\textit{RTE (25, 50)}                                                                                      &            9.88              &          10.5           &          18.0             &       11.25       &         \textbf{7.75}       &           9.88               &           9.5          &           16.75            &       11.75       &      \textbf{7.5}           \\
\hline
\end{tabular}
}
\end{table*}

\subsubsection{Baselines}
We consider five different baselines to compare \model against. The baselines cover both state of the art as well as state of the practise. All algorithms are programmed in Python programming language using Keras version version 2.2.4 and TensorFlow version 1.12.

\begin{itemize}
    \item \textbf{Majority Voting (MV)}: is a basic method which selects from the set of redundant noisy labels $\bm{l}$ the label with the highest consensus.
    \item \textbf{Expectation Maximization (EM)}~\cite{dawid1979maximum}: is an iterative method used to estimate each worker's confusion matrix by maximizing the likelihood of observed labels. The off-diagonal elements represent the probability of mislabeling, the diagonal elements of correct labeling.
    \item \textbf{Bayesian Classifier Combination (BCC)}~\cite{kim2012bayesian}: is an extension of EM. It solves the label aggregation problem by modelling the relationship between the output of multiple classifiers (workers) and the true label.
    \item \textbf{Minimax Entropy (ME)}~\cite{zhou2012learning}: assigns a confusion matrix to  workers, encoding their labeling ability, and a vector to items, encoding their labeling difficulty. The matrix and vector are estimated jointly using a minimax entropy approach.
    \item \textbf{MiniMax Conditional Entropy (MMCE)}~\cite{zhou2014aggregating}: extends ME by assigning confusion matrices also to items intead of a vector. It uses a minimax conditional entropy approach to jointly estimate both worker and item matrices.
    \item \textbf{Label Aware Autoencoders (LAA)}~\cite{yin2017aggregating}: represents the labelling problem via an autoencoder model where the encoder acts as classifier inferring the true label, the decoder reconstructs the input and the inferred labels represent the latent space.
\end{itemize}

\subsubsection{\model Parameters}
As neural network $q$ in \model we use a multi layer perceptron with two hidden layers of size 64 and 32, respectively. The sizes of the input and output layers are given by the number of workers and the number of classes of each dataset, respectively.
We train the network until convergence using marginal loss as stopping criteria.

\subsubsection{Performance Measure}
We use error rate as performance metric in all our experiments. We define the  error rate as the percentage of inferred labels which differ from the true label. Note that the true label is only used to compute the error rate but not to train the label aggregators.



\begin{figure*}[t]
	\centering
	{
	\hfill
	\subfloat[CIFAR-10: chunk size 200, initial set size 1000]{
	    \label{subfig:worker_cif10}
	    \includegraphics[width=0.44\textwidth]{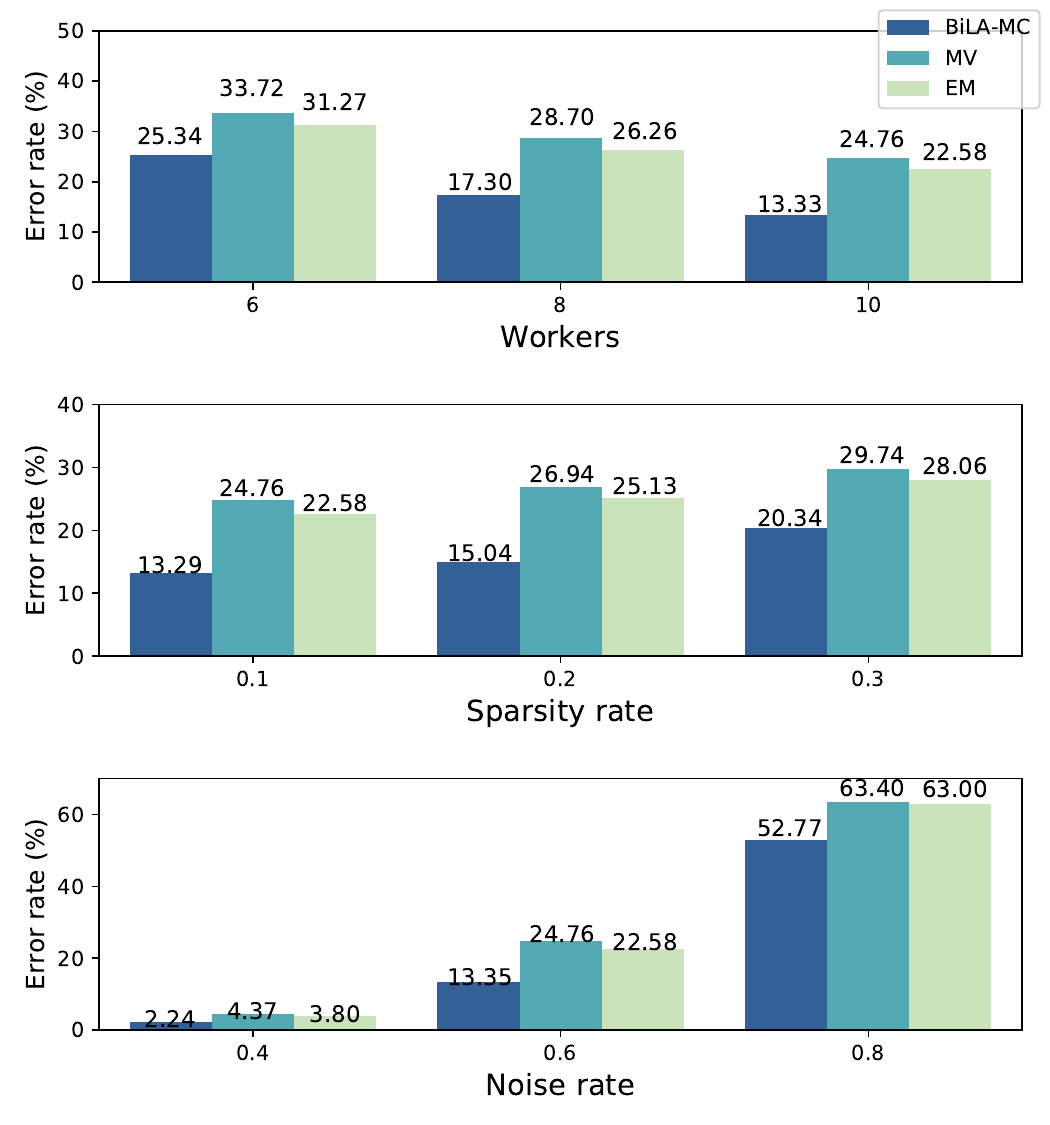} }
	\hfill
	\subfloat[Pendigits: chunk size 200, initial set size 1000]{
	    \label{subfig:worker_pen}
	    \includegraphics[width=0.44\textwidth]{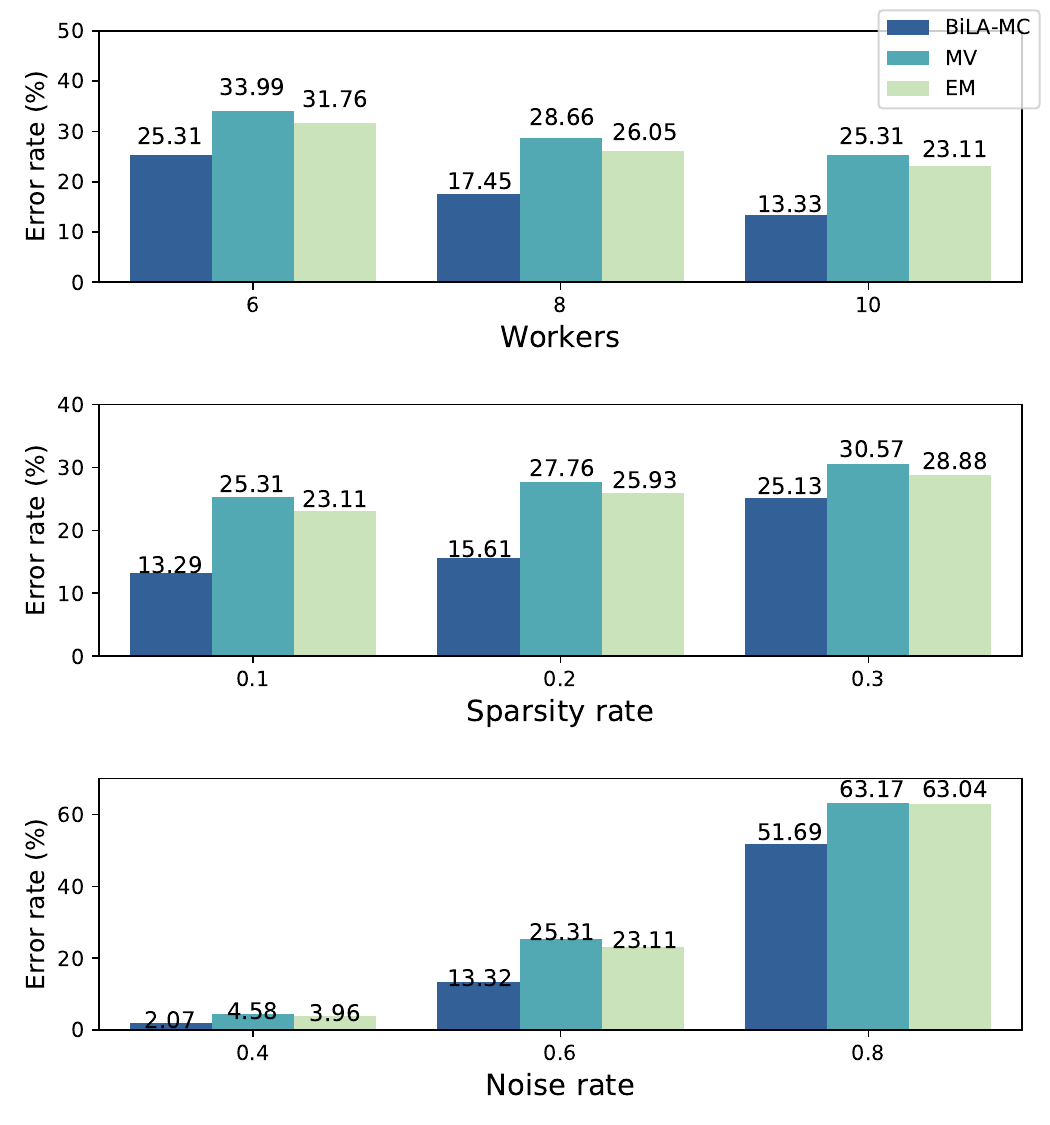}}
	\hfill
	}
	\caption{The effect of the number of workers, label sparsity and noise ratio on \model. The default values for the number of workers, sparsity ratio, and noise ratio are $10$, $0.1$, and $0.6$, respectively. }
	\label{fig:Ana-bila-on-cifar10}
\end{figure*}

\subsection{Results}

We first provide comparative results for online label aggregation, data processed in chunks, showing the superior performance of \model and its robustness to varying chunk sixes. Following we perform a sensitivity analysis of \model on dataset parameters, i.e. number of workers, noise rate and label sparsity, and analyze the optimality of \alg. Finally we conclude with results on offline label aggregation, data processed all at once.

\subsubsection{Online Label Aggregation}

We summarize error rates of \alg and different baselines, across different combinations of datasets and chunk sizes in Table~\ref{tab:online_error-rates}. The small and big chunk size are 200/25 and 500/50 for synthetic/real-world data respectively. Their initial datasets are 1000 and 500 samples for CIFAR-10/Pendigits and Age/RTE, respectively. We initialize all model based approaches by majority voting. Due to the limited number of samples in Adult and Heart dataset, we opt them out from the online evaluation. 

\begin{figure}
	\centering
	{
 	\subfloat[RTE: small chunk size.]{
 	    \label{subfig:RTE:smallchunk}
 	    \includegraphics[width=0.25\textwidth]{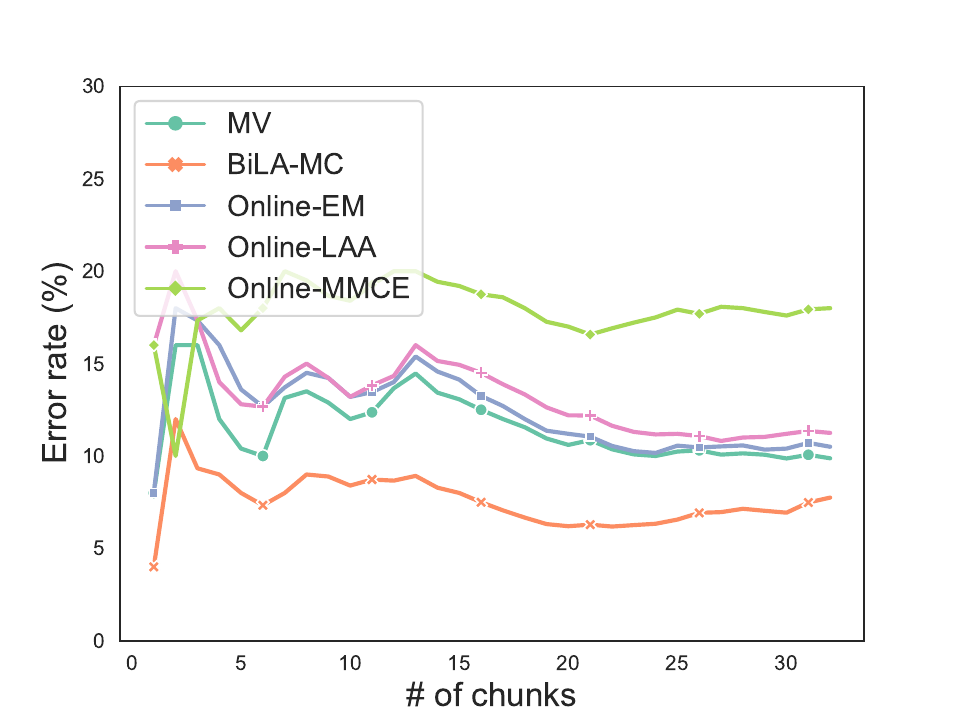}}
    \subfloat[\model: effect of chunk size.]{
	    \label{subfig:bila_bar}
	    \includegraphics[width=0.25\textwidth]{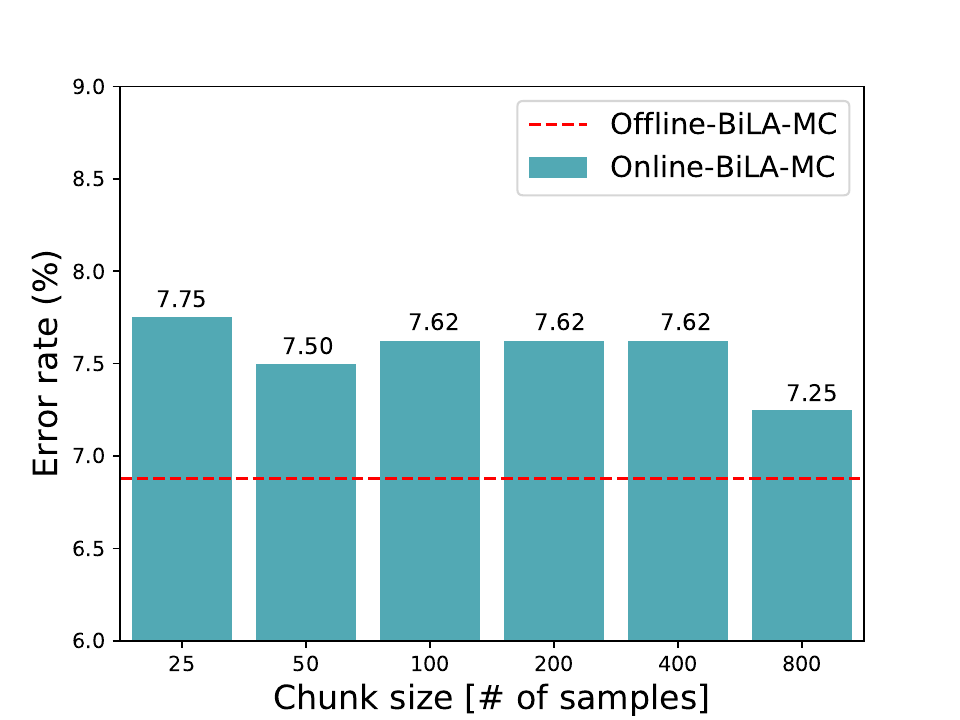}}
	}
	\caption{Online label aggregation of RTE: error rates on \model and the baselines. 
	}
	\label{fig:on-line:aggregation}
\end{figure}

When the chunk size is small, labels from workers are received more fluidly. One can see that \model achieved the lowest error rate. For CIFAR10 and Pendigits, the error rate of \model is 10 percent points lower than the second best algorithm, i.e., online EM. For the Age and RTE datasets, majority voting is the second best algorithm but still has at least 1.5 percent points higher error rates. As MMCE and LAA both have larger number of parameters than EM, their error rates are remarkably high due to insufficient number of samples per chunk for parameterization, especially for MMCE. We further note that smaller chunks not only affect the error rate for EM and MMCE, but also the convergence speed. Due to the small number of samples, it takes MMCE more iterations to converge, compared to big chunk size.

When the chunk size is larger, more items can be aggregated at once, i.e., closer to the offline scenario. \model is still the best algorithm and MMCE comes second, except for Age. The difference from the small chunk size is that now there are sufficient number of samples in a chunk to parametrize the MMCE model. MMCE captures the confusion matrix at the levels of classes, workers and data items. Regarding EM, the error rate drops significantly for bigger chunk size. 

Another observation worth mentioning is the comparison of computational overhead. Due to its simplicity, MV incurs almost no computational overhead. EM algorithm is known to have fast convergence. This is the case observed here. As LAA and \model both employ neural networks, their computational overheads are in the same order.

We further zoom into the error rates over the online aggregation process of the RTE dataset for small chunks (see Figure~\ref{subfig:RTE:smallchunk}). 
When the number of aggregated samples increases,
the error rate first increases and then drops because of the large difference between the size of initial set (i.e., 500 samples) and chunk size (i.e., 25 samples). Overall, \model is able to learn the confusion matrix efficiently from only small chunks of samples and incrementally update the inference model. This is supported by visibly lower error rates across any number of samples processed by the aggregator. 

We also demonstrate the robustness of \model against different chunk sizes or online velocity in Figure~\ref{subfig:bila_bar}. Recalling the motivation examples in Figure~\ref{fig:motivation}, existing label aggregation methods are sensitive to the online velocity, i.e., drastic error rate changes between very big and small chunk sizes. Thanks to incremental updates and stochastic optimization, \model can keep relatively low and constant error rates when encountering different online velocities.

\subsubsection{Sensitivity (robustness) analysis of \alg}
We focus on evaluating the robustness of \model via synthetic redundant noisy labels on two datasets, i.e., CIFAR-10 and Pendigits. Specifically, we evaluate how \model performs against different types of crowd sourcing scenarios, i.e., number of crowd workers, sparsity of labels, and noise rates. The sparsity of labels defines the percentage of missing labels across all items and workers. The noise rate indicates the percentage of wrong labels of all labels collected.

Figure~\ref{fig:Ana-bila-on-cifar10} summarizes such a sensitivity analysis for CIFAR-10 and Pendigits, respectively. We vary one parameter and fix the other two. The default values are 10 workers, sparsity rate of 0.1 and noise rate of 0.6. For the purpose of comparison, we choose the best performing label aggregation methods, i.e., EM and majority voting.

Across all three methods, we can make the following general observations. The error rates decrease with increasing number of workers, and increase with the sparsity and noise rate. In all cases considered, \model always achieves the lowest error rate, followed by EM and then MV. 

Taking a closer look of number of workers, we observe that \model is able to achieve similar error rates, i.e., 25.34\%, as MV but using only 6 instead of 10 workers. As for the robustness against the sparsity rate, EM and MV can better cope with increasing missing labels than \model. Specifically, when the sparsity increases from 0.1 to 0.3, the error rate of \model almost doubles, whereas the error rate of EM only increases by less than 30\%. It appears that \model can be more sensitive to the sparsity than other methods, but the absolute performance is still better. 

Regarding the impact of noise rate, all methods deteriorate drastically when the noise rate is up to 0.8. Actually none of the methods can reach accuracy above 50\%. This is a bottleneck of how label aggregation methods can combat crowd's mistakes. To overcome high percentage of label noise, different solutions may be needed, e.g., a small fraction of ground-truth. When noise ratio is 0.4 and 0.6, we can observe that \model can achieve half of the error rate of the other two methods. 




\subsubsection{Optimality of \alg}

\begin{figure}[t]
\begin{center}
\includegraphics[width=0.85\columnwidth]{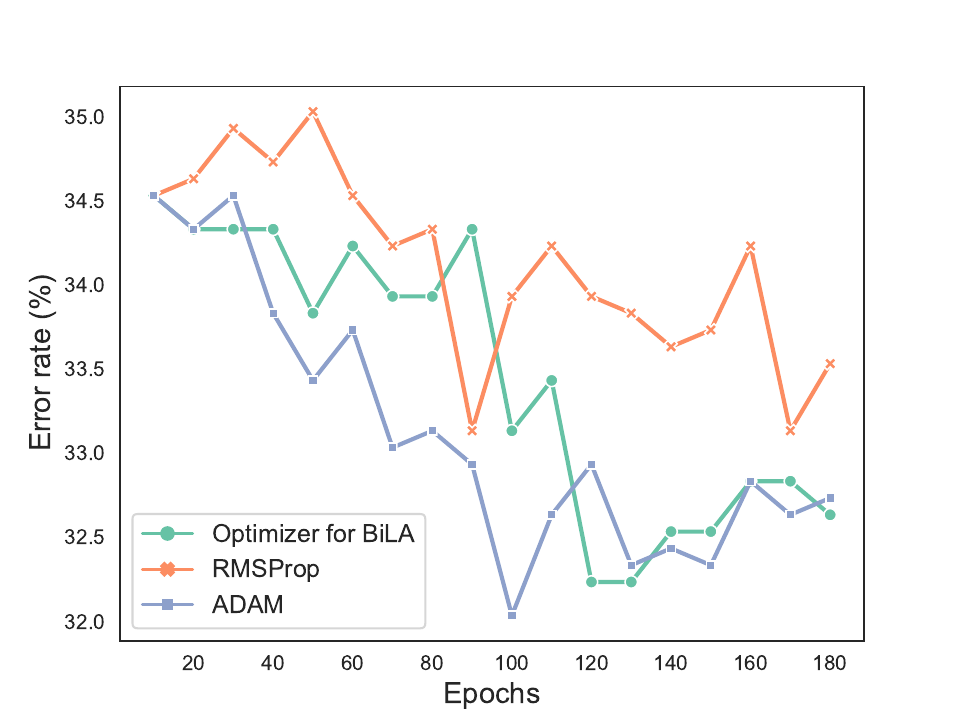}
\end{center}
\caption{Error rates for different optimizers on Age.}
\label{fig:evaluate:optimizers}
\end{figure}

\begin{table*}[t]
\centering
\caption{Offline comparison: error-rates (\%) of label aggregation models}
\label{tab:error-rates}
\begin{tabular}{|c|ccccccc|} 
\hline
\rowcolor{gray!25}
\textbf{Dataset} & \textbf{MV} & \textbf{EM~\cite{dawid1979maximum}} & \textbf{BCC}~\cite{kim2012bayesian} & \textbf{ME\cite{zhou2012learning}} & \textbf{MMCE}~\cite{zhou2014aggregating} & \textbf{LAA~\cite{yin2017aggregating}} & \textbf{\model}  \\ 
\hline
\hline
\textit{Adult}   &    26.43             &        25.48       & 22.81    &  24.33    & 24.33    & 25.86 & \textbf{21.60}                       \\ 
\hline
\textit{RTE}     &     9.88            &         7.5                      & 7.15   &      7.25           &  7.50    &  12.38   &        \textbf{6.88}                \\ 
\hline
\textit{Heart}   &       22.36          &     18.99                          & 18.82     &         16.03        &   16.03   &  13.5   &\textbf{12.66}                        \\ 
\hline
\textit{Age}     &       34.73          &          35.03                    &   33.53  &   32.63   &  32.63   &  34.13  &     \textbf{30.18}                \\
\hline
\end{tabular}
\end{table*}

Next we investigate the optimality of the optimizer used in \model. 
We compare our optimizer against RMSProp~\cite{Tieleman2012} and ADAM~\cite{kingma2014adam}.
Our optimizer is a enhanced version of RMSProp, while ADAM is another common choice.
Figure~\ref{fig:evaluate:optimizers} shows the evolution of the error rate across training epochs. We use the Age dataset with chunk size 25.
We see that our optimizer is faster to converge than RMSProp. Hence we can obtain a higher model performance for the same training effort, i.e. number of epochs.
ADAM is initially slightly faster to converge, but starting at epoch 120 the two optimizers achieve similar model performance.
RMSProp in our implementation and our optimizer both use a clip operator to avoid the issue of gradient explosion which can lead to divergence. ADAM instead is a momentum-based optimizer. Momentum optimizers can be faster to converge than clip-based optimizers but pose the risk of gradient explosion and divergence.
These observations are clearly shown in the results.


\subsubsection{Offline label aggregation}

Finally, we present offline aggregation results in Table~\ref{tab:error-rates}. We include two additional baselines, i.e., BCC~\cite{kim2012bayesian} and ME~\cite{zhou2012learning}. BCC combines the probabilistic models and confusion matrix to infer true labels. ME is the predecessor of MMCE~~\cite{zhou2014aggregating}. Both jointly estimate worker and item latent variables.
Similar to the online results, \model is able to achieve the lowest error rate in all four datasets. The second best policy depends on the dataset. For Adult and RTE, the second best method is BCC that combines probabilistic models and confusion matrix. As for Heart and Age, the second best method is LAA and ME respectively. Though both LAA and \model both use neural networks, \model has a more stable performance due to the guidance of the generating distribution. 

When contrasting the results of small chunk size in Table~\ref{tab:online_error-rates} with results of Table~\ref{tab:error-rates}, we can gauge the impact of online data feeding to different aggregation methods. On the one hand \model has little variation across different online scenarios as it can incrementally update the models chunk by chunk of data. On the other hand, EM and MMCE are observed to have high variability across datasets and online velocity, weakening their applicability for online label aggregation.

\section{Related Work}
Label aggregation is a well-studied subject in crowd sourcing, especially for offline scenarios. Most of existing label aggregation solutions are unsupervised, execpt~\cite{gaunt2016training}. We summarize the related work in accordance with our contributions: (i) the probabilistic inference framework, (ii) confusion matrix aggregation model, and (iii) stochastic models.

\textbf{Probabilistic inference models}.
Probabilistic models are effective to capture how the latent variables, e.g., confusion matrix, affect the likelihood of observed noise labels. 
BCC~\cite{kim2012bayesian} is the very first the probabilistic graphical model for label aggregation. It uses confusion matrix to evaluate workers, and uses Gibbs sampling to perform the parameter estimation. CommunityBCC~\cite{venanzi2014community} and BCCWords~\cite{simpson2015language} are an extension of BCC. Specifically CommunityBCC divides the workers into worker communities. The workers in the same community have similar confusion matrices.  In terms of variational methods, Liu et al.\cite{liu2012variational} propose a model which uses variational inference to approximate the posterior. Recently,~\cite{yin2017aggregating} develops LAA, a label-aware autoencoder. LAA is an unsupervised model composed of a classifier and a reconstructor, both of which are neural networks. ~\cite{li2019exploiting} proposes an offline probabilistic graphical model for label aggregation, including an enhanced variational Bayesian classifier combination with inference based on a mean-field variational method. Orthogonally, Yang et al.~\cite{yang2019scalpel, yang2018leveraging}  apply~ probabilistic inference models applied to jointly distill noisy labels via experts and learning tasks.

Aforementioned studies tailors for offline scenarios where lables of all items is collected at once. And, these methods require to derive a close-form of the generative model's posterior.

\textbf{Confusion matrix}.
Confusion matrix specifies how labels are corrupted from their true class to noisy ones. It can be based on the entre dataset, each worker, and even each content, with  increasing model complexity.
Dawid and Skene~\cite{dawid1979maximum} uses the confusion matrix to describe the expertise and the bias of a worker. They then design an EM algorithm for label aggregation. 
Raykar et al.~\cite{raykar2010learning} uses noisy labels to train their classification model. Their two-coin model is a variation of the confusion matrix.  GLAD~\cite{whitehill2009whose} is a model that can infer the true labels, the expertise of workers, and the difficulty of items at the same time. However, GLAD is applicable for binary labeling tasks.
Furthermore, Zhou et al. \cite{zhou2012learning,zhou2014aggregating} propose the minimax entropy estimator and its extensions. In these model, the authors set a separate probabilistic distribution for each worker-item pair. 

Due to the iterative nature of EM algorithms and minmax entropy, it is not straightforward to extend those methods to construct stochastic optimizer needed for online label aggregation. 

Different from aformentinoed label aggregation methods, DeepAgg \cite{gaunt2016training} is a supervised model based on a deep neural network. The model is trained by a seed dataset which contains noisy labels and the corresponding ground truth labels. DeepAgg can not aggregate incomplete data, where many annotators only labeled a few items. 

\textbf{Stochastic Optimizer} First order stochastic optimization is applied to a wide range of learning problems. RMSProp~\cite{Tieleman2012} and Adam~\cite{kingma2014adam}  are the state-of-the art optimizers. RMSProp updates the model parameters based on the current gradient. It achieves more robust results than stochastic gradient decent because it utilizes element-wise adaptive learning rates to update the model parameters. Adam is a variant of RMSProp. It uses a moving average to estimate the first moment of the gradients and applies the moment to update the model parameters. Often, a clipping operator on the learning rates is used to limit their values during the training and avoid gradient explosion~\cite{mcmahan2010adaptive}. McMahan et al.~\cite{mcmahan2010adaptive} provides the theoretical base for deriving the regret bound of optimizers that use clip operator.


\section{Conclusion}
Motivated by the need of timely and accurately data curation and the avoidance of slow response from crowd workers, we design online label aggregation framework, \alg, maximizing the likelihood of noise labels and inferring unobservable true labels. The core components of \alg are variational Bayesian inference model and a stochastic optimizer for incrementally training on online data subset. The general design of \alg is able to model any generating distribution of labels via  exact computation of posterior probability distribution and  neural networks based approximate distribution. We design a stochastic optimizer that can incrementally minimize the loss function of the variational inference model based on the evidence lower bound. We theoretically prove the convergence bound of the proposed optimizer in terms of parameters of gradient update. We evaluate \alg on both synthetic and real world datasets on various online scenarios. Compared to the state of the art label aggregation algorithms that adopt sliding window update, \alg shows significant and robust error reduction, especially for challenging scenarios with small chunk data set. 


\bibliographystyle{ACM-Reference-Format}
\bibliography{aggregation}

\appendix
\section{Appendix: A Convergence Bound}
\label{Appendix_Bound}

\subsection{Notations and Lemmas}

\textbf{Notations for the proof.} $S_{+}^{d}$ is the set of all positive definite $d \times d$ matrix. $\theta_{,i}^{*}$ is the $i^{th}$ element of $\theta^{*}$. $\theta_{t,i}$ is the $i^{th}$ element of $\theta_{t}$. The operator $\odot$ means element-wise product.

\noindent \textbf{Lemma 1}. \emph{If a function $f: R^{d} \to R$ is convex, then for all $x, y \in R^{d}$, $f(x) - f(y) \le \bigtriangledown f(x)^{T}(x-y)$.}

\noindent \textbf{Lemma 2} (proposed by \cite{mcmahan2010adaptive}). \emph{For any $Q \in S_{+}^{d}$ and closed, bounded convex set $\chi \subset R^{d}$, suppose $u_1 = \min_{\theta \in \chi} \| Q^{1/2}(\theta-z_1) \|_{2}$ and $u_2 = \min_{\theta \in \chi} \| Q^{1/2}(\theta-z_2) \|_{2}$ then we have $\| Q^{1/2}(u_1-u_2) \|_2 \le \| Q^{1/2}(z_1 - z_2) \|_2$.}

\subsection{Proof of Theorem 1}

\noindent \textbf{Theorem 1.} \emph{Let $\{\theta_{t}\}$ be the parameter sequence obtained from our optimizer where $\theta \in R^{d}$. Suppose $\eta_{u}(t) \le R_{\infty}$ and $\frac{t}{\eta_{l}(t)} - \frac{t-1}{\eta_{u}(t-1)} \le B$ for all $t \in [T]$. Assume that $\left\| \theta_n - \theta_m \right\|_{\infty} \le D_{\infty}$ for all $\theta_n, \theta_m \in \chi$ and $\left\| \bigtriangledown f_t(\theta) \right\|_{2} \le G$ for all $t \in [T]$ and $\theta \in \chi$. Our optimizer have the following guarantee of the regret}
\begin{equation*}
    R(T) \le \frac{1}{2}D_{\infty}^{2}\left[ 2dB(\sqrt{T}-1) + \sum_{i=1}^{d} \eta_{1,i}^{-1} \right] + (\sqrt{T}-\frac{1}{2})R_{\infty}G^{2}.
\end{equation*}

\noindent \emph{Proof.} According to Lemma~1, we have 
\begin{equation}
    f_{t}(\theta_t)-f_{t}(\theta^{*}) \le g_{t}^{T}(\theta_t - \theta^{*}) = \left<g_t, \theta_t - \theta^{*} \right>
    \label{eq:f}
\end{equation}

We have the definition $\theta^{*} = \mathop{\arg\min}_{\theta \in \chi} \sum_{t=1}^{T} f_t(\theta)$ mentioned before. According to the update rule shown in Algorithm~\ref{alg:bilamc}, we have $\theta_{t+1} = \min_{\theta \in \chi}\| diag(\eta_{t}^{-1})^{1/2}(\theta - (\theta_{t}-\eta_{t}\odot g_t)) \|_2$. Applying Lemma~2 and setting $u_1 = \theta_{t+1}$ and $u_2 = \theta^{*}$, we have
\begin{align*}
    \| \eta_{t}^{-1/2} &\odot (\theta_{t+1} - \theta^{*}) \|_{2}^{2} \\
    &\le \| \eta_{t}^{-1/2} \odot (\theta_t - \eta_t \odot g_t - \theta^{*}) \|_{2}^{2} \\
    &= \| \eta_{t}^{-1/2} \odot (\theta_t - \theta^{*}) \|_{2}^{2} + \| \eta_{t}^{1/2} \odot g_t \|_{2}^{2} - 2\left<g_t, \theta_t - \theta^{*}\right>
\end{align*}
Rearrange the above inequality, we can have
\begin{align}
    \left< g_t, \theta_t - \theta^{*} \right>  & \le \frac{1}{2}\left[ \| \eta_{t}^{-1/2} \odot (\theta_t - \theta^{*}) \|_{2}^{2} - \| \eta_{t}^{-1/2} \odot (\theta_{t+1}-\theta^{*}) \|_{2}^{2} \right] \notag
    \\ &+ \frac{1}{2} \| \eta_{t}^{1/2} \odot g_t \|_{2}^{2}
    \label{eq:gtheta}
\end{align}
According to (\ref{eq:f}), (\ref{eq:gtheta}) and the definition of the regret $R(T)$ we have
\begin{align}
    R(T) &= \sum_{t=1}^{T}[f_t(\theta_t)-f_t(\theta^{*})] \le \sum_{t=1}^{T} \left < g_t, \theta_t - \theta^{*} \right > \notag \\
    & \le \sum_{t=1}^{T} \frac{1}{2}\left[ \| \eta_{t}^{-1/2} \odot (\theta_t - \theta^{*}) \|_{2}^{2} - \| \eta_{t}^{-1/2} \odot (\theta_{t+1}-\theta^{*}) \|_{2}^{2} \right] \notag \\
    &+ \sum_{t=1}^{T}\frac{1}{2} \| \eta_{t}^{1/2} \odot g_t \|_{2}^{2}
    \label{eq:bound_terms}
\end{align}
We bound $\sum_{t=1}^{T}\frac{1}{2} \| \eta_{t}^{1/2} \odot g_t \|_{2}^{2}$ at first. According to the assumption we have $\|\eta_t\|_{\infty} \le R_{\infty}/\sqrt{t}$ and $\|g_t\|_{2}^{2} \le G^{2}$. Thus we can bound the term as 
\begin{align}
    \sum_{t=1}^{T}\frac{1}{2} \| \eta_{t}^{1/2} \odot g_t \|_{2}^{2} &\le \sum_{t=1}^{T} \frac{R_{\infty}}{2\sqrt{t}} \| g_t \|_{2}^{2} \le \frac{R_{\infty}G^{2}}{2}\sum_{t=1}^{T} \frac{1}{\sqrt{t}}  \notag \\
    & \le (\sqrt{T}-\frac{1}{2})R_{\infty}G^{2},
    \label{eq:bound_term1}
\end{align}
where $\sum_{t=1}^{T}\frac{1}{\sqrt{t}} \le 2\sqrt{T}-1$. Now, we bound the another term in (\ref{eq:bound_terms}).

\begin{align}
    &\sum_{t=1}^{T} \frac{1}{2}\left[ \| \eta_{t}^{-1/2} \odot (\theta_t - \theta^{*}) \|_{2}^{2} - \| \eta_{t}^{-1/2} \odot (\theta_{t+1}-\theta^{*}) \|_{2}^{2} \right] \notag \\
    &= \sum_{i = 1}^{d}\sum_{t=1}^{T}\frac{1}{2}\left[ \eta_{t,i}^{-1}(\theta_{t,i}-\theta_{,i}^{*})^{2}-\eta_{t,i}^{-1}(\theta_{t+1,i}-\theta_{,i}^{*})^{2} \right] \notag \\
    & \le \sum_{i=1}^{d} \left[ \frac{1}{2} \eta_{1,i}^{-1}(\theta_{1,i}-\theta_{,i}^{*})^{2} + \sum_{t=2}^{T}\frac{1}{2}(\eta_{t,i}^{-1}-\eta_{t-1,i}^{-1})(\theta_{t,i}-\theta_{,i}^{*})^{2} \right] \notag \\
    & \le \sum_{i=1}^{d} \left[ \frac{1}{2} \eta_{1,i}^{-1}(\theta_{1,i}-\theta_{,i}^{*})^{2} + \sum_{t=2}^{T}\frac{1}{2}\left[ \frac{\sqrt{t}}{\eta_{l}(t)}-\frac{\sqrt{t-1}}{\eta_{u}(t-1)} \right](\theta_{t,i}-\theta_{,i}^{*})^{2} \right] \notag \\
    & \le \frac{1}{2} D_{\infty}^{2} \sum_{i=1}^{d} \left[\eta_{1,i}^{-1} + \sum_{t=2}^{T}\frac{1}{\sqrt{t}}\left[ \frac{t}{\eta_{l}(t)}-\frac{t-1}{\eta_{u}(t-1)} \right] \right] \notag \\
    & \le \frac{1}{2} D_{\infty}^{2} \sum_{i=1}^{d} \left[ \eta_{1,i}^{-1} + B\sum_{t=2}^{T}\frac{1}{\sqrt{t}} \right] \notag \\
    & \le \frac{1}{2} D_{\infty}^{2} \left[ 2dB(\sqrt{T}-1) + \sum_{i=1}^{d}\eta_{1,i}^{-1} \right]
    \label{eq:bound_term2}
\end{align}
In the second inequality we use the inequation $\eta_{l}(t) \le \eta_{t,i} \le \eta_{u}(t)$ which can be obtained by the clip operator. In the third inequality we applied the bound $D_{\infty}$. In the fourth we applied the assumption $\frac{t}{\eta_{l}(t)} - \frac{t-1}{\eta_{u}(t-1)} \le B$. Then according to (\ref{eq:bound_terms}), (\ref{eq:bound_term1}) and (\ref{eq:bound_term2}) we have the following regret bound
\begin{equation*}
    R(T) \le \frac{1}{2}D_{\infty}^{2}\left[ 2dB(\sqrt{T}-1) + \sum_{i=1}^{d} \eta_{1,i}^{-1} \right] + (\sqrt{T}-\frac{1}{2})R_{\infty}G^{2}.
\end{equation*}

\section{Appendix: One Binary label aggregation model}

In this section we show the definition of a binary label aggregation model. This model is called \modelapx. In \modelapx, $q$ is a MLP. It inputs an instance $\bm{l}$ and outputs a distribution $q_{\bm{\alpha}}(t|\bm{l})$, where $\bm{\alpha}$ denotes the network parameters. 

Next, we define a generative model $p$ to describe the generation of the observed noisy labels. As shown in (\ref{eq:klcalc}) and (\ref{eq:ecalc}), in order to compute the loss function and its gradient, we need to define $p_{\bm{\beta}}(\bm{l}|y)$ and $p_{\bm{\beta}}(y)$. In NN-WA, we only consider binary labeling tasks. For each $c \in \{1,2\}$, the ability of each worker $k$ is represented by a single parameter $\lambda_{ck} \in (-\infty,+\infty)$. We assume that worker $k$ labels each item $i$ correctly with the probability
\begin{equation}
p_{\bm{\beta}}(l_{ik} = c|y_i = c) = \frac{1}{1+\mathrm{e}^{-\lambda_{ck}}},
\label{eq:workercorr}
\end{equation}
According to this assumption, we have $\lim_{\lambda_{ck} \to +\infty} p_{\bm{\beta}}(l_{ik} = c|y_i=c) = 1,$ $\lim_{\lambda_{ck} \to -\infty} p_{\bm{\beta}}(l_{ik} = c|y_i=c) = 0,$ and $\lim_{\lambda_{ck} \to 0} p_{\bm{\beta}}(l_{ik} = c|y_i=c) = 0.5.$ We can see that the higher the ability of worker $k$ is, the higher the likelihood for him or her to label the item correctly. When $\lambda_k = 0$, he or she just randomly chooses one class. According to (\ref{eq:workercorr}), the conditional distributions that generated instances are defined as
\begin{equation}
p_{\bm{\beta}}(\bm{l}_i|c) = \prod_{k \in \bm{S}_i}\left(\frac{1}{1+\mathrm{e}^{-\lambda_{ck}}}\right)^{\mathbbm{I}(l_{ik}=c)}
\left(\frac{\mathrm{e}^{-\lambda_{ck}}}{1+\mathrm{e}^{-\lambda_{ck}}}\right)^{\mathbbm{I}(l_{ik}\neq c)},
\end{equation}
where $\bm{S}_i$ is a set of workers who have labeled item $i$. In this model, the prior distribution $p_{\bm{\beta}}(y)$ is fixed during the training process. It can also be estimated by Equation~(\ref{eq:priorest}). The optimization goal of \modelapx also takes the form as Equation~(\ref{eq:minilossprac}). \modelapx can also be applied online like \model.

\end{document}